# Highly accurate model for prediction of lung nodule malignancy with CT scans

Jason Causey[1,7†], Junyu Zhang[2†], Shiqian Ma[3], Bo Jiang[4], Jake Qualls[1,7], David G. Politte[5], Fred Prior[6*], Shuzhong Zhang[2*] and Xiuzhen Huang[1,7*]

[1]Department of Computer Science, Arkansas State University, Jonesboro, Arkansas 72467, United States of America
[2]Department of Industrial and Systems Engineering, University of Minnesota, Minneapolis, Minnesota 55455, United States of America
[3]Department of Systems Engineering and Engineering Management, The Chinese University of Hong Kong, Shatin, N.T., Hong Kong
[4]Research Center for Management Science and Data Analytics, School of Information Management and Engineering, Shanghai University of Finance and Economics, Shanghai 200433, China
[5]Mallinckrodt Institute of Radiology, Washington University, St. Louis, Missouri 63110, United States of America
[6]Department of Biomedical Informatics, University of Arkansas for Medical Sciences, Little Rock, Arkansas 72205, United States of America
[7]The UALR/UAMS Joint Graduate Program in Bioinformatics, Little Rock, Arkansas 72204, United States of America

† The first two authors are considered as joint first authors.
*Corresponding authors

**Abstract.** Computed tomography (CT) examinations are commonly used to predict lung nodule malignancy in patients, which are shown to improve noninvasive early diagnosis of lung cancer. It remains challenging for computational approaches to achieve performance comparable to experienced radiologists. Here we present NoduleX, a systematic approach to predict lung nodule malignancy from CT data, based on deep learning convolutional neural networks (CNN). For training and validation, we analyze >1000 lung nodules in images from the LIDC/IDRI cohort. All nodules were identified and classified by four experienced thoracic radiologists who participated in the LIDC project. NoduleX achieves high accuracy for nodule malignancy classification, with an AUC of ~0.99. This is commensurate with the analysis of the dataset by experienced radiologists. Our approach, NoduleX, provides an effective framework for highly accurate nodule malignancy prediction with the model trained on a large patient population. Our results are replicable with software available at http://bioinformatics.astate.edu/NoduleX.

Lung cancer is the leading cause of cancer related death worldwide [Siegel et al. 2014]. Early diagnosis of lung cancer is key to reducing mortality. Screening computed tomography (CT) examinations have been shown to greatly improve noninvasive early diagnosis of lung cancer in at risk patients [Atwater et al. 2016; Aberle 2016]. The National Lung Screening Trial (NLST) has shown that low dose computed tomography screening could reduce patient mortality by 20% compared with conventional chest radiographs [Aberle et al. 2011]. The US Preventive Services Task Force (USPSTF) recommends annual screening for lung cancer with computed tomography for high risk patients. Our ability to conduct lung nodule malignancy classification from clinical CT data has important clinical impacts. However, for current clinical evaluation, radiologist's interpretation is a tedious process, which is an impediment of clinical throughput, and highly subjective. It would be ideal to develop computational approaches, with performance comparable to radiologists, in order to aid or ease the burden of radiologists, and also potentially identify and make use of nodule image features that may be missed by even very experienced thoracic radiologists.

There are several substantial challenges for computational approach development with currently available CT datasets: (i) Computational approaches need to provide sensitivity to nodule features as well as robustness to noise and other artifacts introduced during data collection. (ii) Datasets need to be large enough, consistent, and ensure the integrity of expert-defined ground "truth" for approach development, training, and testing [Armato2007]. The definition of "truth" and other statistical issues regarding database design for training computer-aided diagnosis tools are discussed in [Dodd et al. 2004]. The development of a successful computational approach should take these challenges into consideration.

There are two general categories of computational strategies recently developed for lung nodule malignancy prediction from CT images: (i) Radiomics approaches based on radiological quantitative image features (QIF). (ii) Deep learning approaches such as those based on convolutional neural networks (CNN). Radiomics approaches [Liu et al. 2016, Aerts et al. 2014, Gierada et al. 2016; Hawkins et al. 2016; Reeves et al. 2016; Dilger et al. 2015; Firmino et al. 2016; Wang et al. 2016] usually build the prediction model based on the extracted two dimensional (2D) or three dimensional (3D) radiological quantitative image features of lung nodules based on prior knowledge of what features and characteristics are significant. Radiomics approaches have been developed using publicly available datasets, such as The Lung Image Database Consortium (LIDC/IDRI) and the National Lung Screen Trial (NLST) [Clark et al. 2013], or using proprietary datasets, which are frequently small but may be confirmed via pathology based on biopsies or surgical resections. Deep learning convolutional neural network (CNN) based approaches are very promising with the availability of CT scans from large cohorts. Many recent efforts are devoted to nodule classification approaches using convolutional neural networks [Greenspan et al. 2016; Li et al. 2016; Shen et al. 2017]. For example, [Kumar et al. 2015; Golan2016] have used data from the LIDC/IDRI cohort to train and test their models for classifying lung nodules.

There are several differences between the radiomics approach and the CNN based approach, which need to be taken into consideration in order to develop successful models based on either approach or potentially to integrate of the two approaches. The two approaches require different input information for nodule malignancy prediction. Radiomics approaches need proper segmentations of the nodules from radiologists or from segmentation algorithms, and then need quantitative image feature extraction. CNN approaches do not necessarily require segmentation of the nodules and can perform prediction with one marked point per nodule after the prediction model is trained. While radiomics approaches are based on radiological quantitative image features, the features learned by deep convolutional neural network approaches may be visualized as mysterious "deep dreams" [Mordvintsev et al. 2015]. Emerging as the leading machine-learning approach in the imaging domain, deep learning CNN approaches usually require a much larger training dataset, compared with radiomics approaches. Once trained, the CNN models are can be more efficient for nodule malignancy prediction, compared with the models based on radiomics approaches, since the prediction can be made directly from the image without the need for a quantitative feature extraction step prior to classification.

While much progress has been made in the development of models based on either radiomics or deep learning CNNs for lung nodule malignancy classification, it remains challenging for computational approaches to achieve performance comparable to experienced radiologists. Here we present NoduleX, a novel systematic approach for lung nodule malignancy classification from clinical CT scans. The approach is based on deep learning convolutional neural network (CNN) features, and it can also integrate the information of radiological quantitative image features (QIF), if available. For training and validation with lung nodules in images from the LIDC/IDRI cohort, NoduleX achieves high accuracy for nodule malignancy classification, commensurate with the analysis of the dataset by experienced radiologists.

**RESULTS**
**NoduleX: an approach for nodule malignancy classification**
NoduleX is a novel, systematics approach for lung cancer nodule malignancy classification using clinical CT scans. The approach is based on deep learning convolutional neural network (CNN) features. The general framework of NoduleX is illustrated in **Figure 1**. We describe the details in the Methods Section.

We followed a rigorous process for training and testing the model using CT data from the LIDC/IDRI participant cohort [Clark et al. 2013] (see supplemental material.). In the LIDC study, four experienced thoracic radiologists reviewed each of the 1018 CT cases in the LIDC/IDRI cohort and marked lesions belonging to one of three categories ("nodule > or = 3 mm," "nodule < 3 mm," and "non-nodule > or = 3 mm"). The lesions of nodules ≥ 3 mm have a greater probability of malignancy than lesions in the other two categories. The malignancy rating (1, 2, 3, 4, and 5) of the nodules ≥ 3 mm from the four experienced radiologists are described in detail in the two publications of the LIDC/IDRI cohort [McNitt-Gray et al. 2007; Armato et al. 2011]. This malignancy score/rating of this cohort was also discussed in other related recent studies of the cohort such as [Tan et al. 2010; Ypsilantis and Montana, 2016]. Of the cohort, there is diagnostic data for 157 of the 1018 patients, which were obtained by performing biopsy, surgical resection, and progression or reviewing the radiological images to show 2 years of nodule state at two levels. The nodule level diagnosis of the 157 patients is: unknown, benign, malignant (primary lung cancer), and malignant (metastatic).

**NoduleX has consistent performance with high accuracy**
We processed 1065 nodules with different malignancy scores from 1-5 (with score 1 meaning highly unlikely to be malignant, score 2 or 3 indeterminate, score 4 moderately likely to be malignant, and score 5 highly likely to be malignant). The corresponding sets are denoted as S1, S2, S3, S4 and S5, respectively. We tested two designs: S1 versus S45, and S12 versus S45. For each design, the data were grouped into completely independent training and validation sets, with 80% for training and 20% for validation. Both the training and the validation sets was balanced to contain an equal number of two classes of nodules as "malignant" and "benign" nodules. **Figure 2(a)** shows an example of two patients' CT scan slices of nodules with malignancy score 1 and score 5 respectively, which are reviewed by the LIDC/IDRI cohort experienced thoracic radiologists.

For the design of S1 versus S45, the best model on the validation set has an area under the receiver operating characteristic curve (AUC) of 0.974 (acc = 91.3%, sen = 88.5%, spc = 94.2%). The model performance is further improved when combined with the identified radiomic quantitative image features (QIF), with an AUC of 0.993 (acc = 95.2%, sen = 94.2%, spc = 96.2%). For the design of S12 versus S45, the best model on the validation set has an AUC of 0.938 (acc = 87.9%, sen = 87.9%, spc = 87.9%). When combined with the QIF features, the model performance is further improved with an AUC of 0.971 (acc = 93.2%, sen = 87.9%, spc = 98.5%). **Figure 3(a) and (b)** show the receiver operating characteristic curve (ROC) of two CNN models, two CNN with combined QIF features to predict malignancy on the validation set. Please refer to **Table 1** for each model, the AUC, accuracy sensitivity specificity, for the validation set.

We also investigated the performance of NoduleX on prediction of nodules versus non-nodules. We used 1067 nodules with malignancy scores of 1-5 and 1056 non-nodule points, reviewed by the four radiologists of the LIDC/IDRI cohort; see **Figure 2(b)** for an example of two patients' CT scan slices of a nodule with malignancy score 3 and a non-nodule point respectively. Of the ~2000 instances of nodule and non-nodule data, 80% of the data are used for training and 20% for validation. Both the training and the validation sets were balanced to contain an equal number of nodules and non-nodule points. We tested the models' performance on the independent validation set. For prediction of nodules versus non-nodules for the independent validation test, the model AUC is 0.949 (acc = 89.9%, sen = 87.7%, spc = 92.0%) and the model AUC is 0.984 (acc = 94.6%, sen = 94.8%, spc = 94.3%) when combined with the QIF features. **Figure 4** shows the receiver operating characteristic curve (ROC) of two CNN models, two CNN with combined QIF features to predict malignancy on the validation set. Please refer to **Table 1** for the AUC, accuracy sensitivity specificity, for the validation set, compared with the two designs: S1 versus S45, and S12 versus S45. In contrast to other models developed in the literature using the LIDC/IDRI dataset with AUC at the range of ~0.8, our model achieved significantly improved performance accuracy, with matched performance as experienced radiologists for the LIDC/IDRI cohort.

**Radiomics approach for nodule malignancy classification of the LIDC/IDRI cohort**
The radiomics features crafted to capture visual cues that radiologists identified, in combination with radiologist segmentations tended to agree closely with the radiologist's assigned classifications, even when a very small number of samples were used for training. Our testing, described below and quantified in **Table 2**, revealed that radiomic quantitative image features are able to describe the differences in nodules that are identified by experienced radiologists as belonging to different classes (e.g., "Highly unlikely for cancer" and "Highly suspicious for cancer"). We show that even when small sample sizes are used for training, good separation can be achieved by the radiomics approach.

To establish a lower limit on how well the radiomics features, with a Random Forest (RF) model, could classify the nodules, we conducted the following tests:

S1 vs S45 RF separation using 1+ and 1- training set:
1 positive (S45) nodule and 1 negative (S1) nodule chosen at random from the full set of 520 nodules were used for training a RF classifier, with the remaining 518 as the "test set"; the test was repeated 200 times, and the results averaged: mean AUC = 0.91; mean acc = 81%.

S12 vs S45 RF separation using 1+ and 1- training set:

1 positive (S45) nodule and 1 negative (S12) nodule chosen at random from the full set of 664 nodules were used for training a RF classifier, with the remaining 662 as the "test set"; the test was repeated 200 times, and the results averaged: mean AUC = 0.86; mean acc = 75%.

Additionally, to establish a baseline for the separation difficulty, a logistic regression model (LM) was trained only on a size metric (square root of largest cross-sectional area). For S0 vs S1-5, i.e., non-nodule vs nodule, the separation test was also conducted.

Please refer to **Table 2** for the comparison of the quantitative image feature (QIF) models with the baseline model for the designs of S1 vs S45, S12 vs S45, and S0 vs S1-5.

**DISCUSSION**
We present NoduleX, an effective framework for lung nodule malignancy prediction from patients' CT scans. For training and validation, we performed analysis of the nodules of the LIDC/IDRI cohort and we found that NoduleX can achieve ~0.99 AUC on the independent validation test. Unlike existing models usually with moderate accuracy levels, our testing results demonstrate that NoduleX can achieve high prediction accuracy, commensurate with the reviews of the cohort from experienced radiologists. To this end, NoduleX leverages the following several properties: (i) The model is developed based on deep learning CNN features, which can incorporate "window" normalization based on the Hounsfield unit for the input CT scan image, and apply enhanced classifiers such as Random Forest, LogitBoosting or AdaBoosting. (ii) The training and validation sets are carefully designed from the LIDC/IDRI cohort to avoid potential statistical biases or issues. The model is trained using the nodules with different malignancy score of the LIDC/IDRI cohort and validated using completely independent validation sets developed from the cohort. (iii) The model can integrate deep learning CNN features (CNN feature expression) with radiological quantitative image features (radiomic expression), if the information of the nodules is available. The features of the CNN model and the QIF model are different, and the consistency of the testing results from the two kinds of models confirms the performance of each model.

There are in general three ways of using the data of the LIDC/IDRI cohort for the study related to classifying nodule as two classes of "benign" and "malignant":

(i) Using the set of nodules with the malignancy score/ratings (1, 2, 3, 4, and 5), reviewed by the experienced radiologists of the cohort. We treat the malignancy of nodules as a binary classification problem for "malignant" versus "benign" by thresholding the radiologist-assigned malignancy values so that malignancy values below 3 (i.e., 1 and 2) are categorized as benign and values above 3 (i.e., 4 and 5) are categorized as malignant. Recent published models for the problem of classifying S12 (considered as "benign") versus the S45 (considered as "malignant") include the models developed in [Silva et al. 2016; Cheng et al. 2016; Hancock and Magnan, 2016]. In [Silva et al. 2016], Taxonomic indexes and phylogenetic trees were used as texture descriptors, and a Support Vector Machine was used for classification. The proposed method shows promising results for accurate diagnosis of benign and malignant lung tumors, achieving an accuracy of 88.44%, sensitivity of 84.22%, specificity of 90.06% and area under the ROC curve of 0.8714.

(ii) Using only the set of nodules of the patients with diagnosis data. As in [Shewaye and Mekonnen, 2016], only those nodules with benign and malignant labels are considered. They used 52 subjects with malignant nodules and 21 subjects with benign nodules with a total of 458 and 107 individual lung nodules respectively. The obtained results are very encouraging, correctly classifying 82% of malignant and 93% of benign nodules on unseen test data at best.

Similarly in [Kumar et al. 2017] the authors used the clinically provided pathologically-proven data as ground truth, and obtained an average accuracy of 77.52% with a sensitivity of 79.06% and specificity of 76.11%.

(iii) Using both sets. The authors in [Kumar et al. 2015] used the ratings from diagnostic data as the ground truth for training the classification system and evaluating the results instead of using the radiologists provided ratings. In [Kumar2015], the author reported the trained model obtained an overall accuracy of 75% for classifying whether the nodule is benign or malignant.

Our rationale for the Nodule versus Non-Nodule classification task was to use the full LIDC/IDRI cohort to address the task first presented in the LUNA16 grand challenge of *LUng Nodule Analysi*s [Setio et al. 2016].

There are several common pitfalls when evaluating and testing a classification model for lung nodule malignancy: (i) not using a completely independent validation set; (ii) not considering the nodules of the same patient to be completely separated into the training set or the validation set; (iii) not testing the model without size-related image features in order to remove the potential bias introduced by nodule size. (iv) not reporting the classification testing results with the complete information for AUC, as well as for accuracy, sensitivity, and specificity. For example, if an unbalanced number of "benign" and "malignant" nodules for the validation set is used, the testing result may have a very high accuracy rate but with a very low specificity.

As is pointed out in [Shewaye and Mekonnen, 2016], since most of the research groups report experimental results using their own proprietary dataset that is not publicly available or a different subset of a publicly available dataset, a direct absolute comparison of their statistics performance is not possible. Therefore, it is difficult to cross-validate the developed computational approaches with a completely different dataset. Please refer to a summary of related works in "benign" vs "malignant" lung nodule classification in [Shewaye and Mekonnen, 2016].

Evaluating the performance of NoduleX introduces several challenges: (i) The number of patient CT scans in the LIDC/IDRI cohort is not large enough for training very sophisticated CNN classification models. For example, with a large number of images, the learning model for face recognition of Facebook was trained on four million images and it is said that the model reached an accuracy level even higher than the FBI current system for face recognition. (ii) For training and validating we use the LIDC/IDRI dataset. It may not be directly applicable to other datasets with very different CT scan image quality or different definitions for ground truth classifications. If two datasets are of very different quality or if nodule ground truth labels are not consistent, the computational model trained on one dataset will need to be re-trained in order to work for the other dataset. We may consider for future research work on transfer learning and testing on other screening and diagnostic data of CT images. There are some recent works on transfer learning for survival prediction analysis such as in [Paul et al. 2016]. (iii) We cannot make a direct performance comparison of our model with previously developed models. Although the models developed in [Cheng et al. 2016] were trained and tested on the same problem of classifying S12 vs S45, the nodule sets were chosen differently from the LIDC/IDRI cohort. (iv) The nodules of the LIDC/IDRI cohort with malignancy scores of 4 or 5 were not necessarily confirmed through biopsies or surgical resection as real malignant nodules. We plan to test and cross-validate the approach with other datasets where diagnostic truth has been established.

Interestingly, our testing also reveals the problem of lung nodule malignancy prediction with the available nodule classification from experienced radiologists as ground truth of the LIDC/IDRI cohort, which has been widely studied in the area, is actually a relatively "easy" radiomics problem. Our testing shows that radiomics features are able to describe the differences in nodules that are identified by a human expert as belonging to very different classes. Our testing results demonstate that even when small sample sizes are used for training, a radiomics prediction model can achieve reasonably good separation. This lends confidence that the radiomic quantitative features are robustly representing information that a human would use to classify the nodules. Of course, it leaves open the question of whether these nodule-level classifications are a good predictor of patient outcome – additional research on a dataset where outcome is included will be necessary to address this.

With NoduleX as a systematic framework, the model can be re-trained with other large datasets of CT scans, and we anticipate it can achieve similar high accuracy on other datasets. It can be re-trained when new data becomes available and can continue to learn from the increasing knowledge of radiologists as well as be further trained with more available CT scans from a larger population. This research will help open a new path for developing effective computational approaches based on deep CNN features in medical imaging, which may have a clinical impact as images are routinely collected in clinical practice, for disease diagnosis, prognosis and treatment.

NoduleX makes a substantial step towards addressing the challenge that computational prediction for lung nodule malignancy with patients CT scans can have matched performance as the reviews of experienced thoracic radiologists in current clinical practice. We performed analysis of >1000 nodules of the CT scans of 1018 patients of the LIDC/IDRI cohort. All these nodules were identified and classified by four experienced thoracic radiologists who participated in the LIDC project. Our test shows that NoduleX, when provided with CT data and a point locating the nodule, achieves an area under the receiver operating characteristic curve (AUC) of 0.97. The performance is further improved when combined with quantitative image features (QIF), resulting in an AUC of 0.99. Compared with previous work for this problem based on QIF or CNN in the literature, NoduleX achieved significant performance accuracy, commensurate with the reviews of the LIDC/IDRI cohort by the experienced radiologists. As the recent great advancements in deep learning approaches for voice recognition and face recognition with large available datasets, we expect advancements in computational learning approaches in biomedical imaging for lung cancer early detection and diagnosis.

## METHODS
Methods and any associated references are included in **Supplementary** information.

**NoduleX for nodule malignancy classification.** Here we describe the details of NoduleX, for lung nodule malignancy classification using clinical CT scans. The approach is based on deep learning convolutional neural network (CNN) features. The general framework of NoduleX is illustrated in **Figure 1** and the CNN network layout for the two networks reported here is illustrated in **Figure 5**.

**Nodule Selection.** The nodule annotations provided in XML format with the LIDC-IDRI CT image dataset were used to produce a consensus list of nodules for each patient such that the nodules in the consensus list had no overlap and the malignancy rating assigned was the average of the malignancy ratings assigned by the radiologists who annotated the nodule, rounded to the nearest integer. One nodule was then chosen at random from each patient in the study who had at least one lesion marked as a "nodule ≥ 3mm" by at least one radiologist. This random selection produced 44 nodules rated "malignancy = 1" and 270 nodules rated "malignancy = 4" or "malignancy = 5"; to balance the size of the positive and negative classes, an additional 206 nodules rated "malignancy = 1" were selected by choosing all unique nodules rated "malignancy = 1" from patients for whom a nodule rated "malignancy = 4" or "malignancy = 5" was not selected in the random selection phase. (Note: some of these patients do have "malignancy = 4 or 5" that were not selected by the initial random sampling procedure.) This resulted 250 nodules rated "malignancy = 1" and 270 nodules rated "malignancy = 4" or "malignancy = 5".

**CNN Input Volume Extraction.** Input to the CNN consists of a small 3-D volume measuring either 21 pixels × 2 pixels × 5 slices, or 47 pixels × 47 pixels × 5 slices, depending on the CNN architecture. These volumes were extracted from the full CT scan by selecting the 3D region centered around the nodule's center of mass (centroid), as determined by consensus of the segmentations from all radiologists who segmented the nodule (the average centroid among all segmentations was used). These rectangular volumes were saved along with the consensus malignancy rating of the nodule, the minimum and maximum pixel intensities of the scan, and a class identifier in a file in HDF5 format. Separate "train" and "validation" sets were created as separate HDF5 files in this way; the "train" set contained 442 nodules and the "validation" set contained 110 nodules; all nodules in the "train" and "validation" sets were matched to corresponding nodules in the QIF "train" and "validation" sets.

**CNN Training.** The CNN was trained by further dividing the "train" set into a training group consisting of 80% of the included nodules and a testing group containing the other 20%; this division was performed at random at the start of each training run. Training continued for 200-400 epochs, and the batch size was 64. At the end of each epoch a checkpoint of the model weights was saved if the model loss was improved. The final model weights as well as the three checkpoints with the highest accuracy on the testing portion of training data were retained. To reduce overfitting, automatic data augmentation was performed in which each input image volume was randomly shifted up to 30% in both the X and Y directions, and randomly rotated between 0 and 180 degrees. After training, the retained model weights were evaluated against the separate "validation" set; results for the best 21×21×5 and 47×47×5 model are reported. The *Keras* software package with the *Theano* backend was used for implementation, training, and testing of the CNN model.

**CNN Feature Extraction.** After training is completed, features are extracted from the validation set by providing the set of nodule volumes to be evaluated to the CNN network in prediction mode. Output values are captured from the fully-connected layer just prior to the 2-class classification layer (i.e. the second-to-last layer). These values form a feature vector for each nodule, and are aggregated into an output CSV file for further processing in combination with QIF features.

**NoduleX Classification**. Nodule classification was performed using two different CNN models (the CNN21 and CNN47 models) and both with and without QIF features. For classification without QIF features, the CNN model's own output softmax classifier was used for class prediction on all nodules in the validation set. When QIF features are used, a vector representing the 50 QIF features is concatenated with the feature vector produced by the CNN as described above, producing a feature vector with 250 features. This combined feature vector is passed as input to a Random Forest classifier model, which itself must be trained on the training set (as described above). The trained Random Forest is then evaluated on the same validation set as the CNN-only model for comparison. The *randomForest* package [Liaw and Wiener, 2002] in R was used for this purpose, with the *ntrees* parameter set to 1000 and defaults for other parameters.

**Description of the datasets**.
The LIDC/IDRI datasets contains the CT scans of 1018 patients/cases, and some patients may have more than one nodule. These CT scans were reviewed by four experienced thoracic radiologists. The radiologists annotated each scan by marking regions of interest in three classes: "nodule ≥ 3mm," "nodule < 3mm," and "non-nodule." Each nodule in the "nodule ≥ 3mm" class was then given a malignancy score and a detailed segmentation. "Non-nodule" and "nodule < 3mm" regions were noted by position in the scan only. The malignancy scores were defined as follows: 1 "Highly Unlikely for Cancer," 2 "Moderately Unlikely for Cancer," 3 "Indeterminate Likelihood," 4 "Moderately Suspicious for Cancer," 5 "Highly Suspicious for Cancer."

**Dataset S1 vs S45.** This dataset consists of 520 nodules with malignancy ratings of 1, 4, or 5 as determined by consensus of radiologist ratings from the metadata provided with the LIDC/IDRI cohort. Nodules with malignancy = 1 were designated as the "negative" class (S1) and nodules with malignancy = 4 or 5 were designated as the "positive" class (S45). There were 250 S1 nodules and 270 S45 nodules in total.
**Dataset S12 vs S45.** This dataset consists of 664 nodules with malignancy ratings of 1, 2, 4, or 5 as determined by consensus of LIDC/IDRI radiologist ratings. Nodules with malignancy = 1 or 2 were designated as the "negative" (S12) class, and nodules with malignancy = 4 or 5 were designated as the "positive" (S45) class. There were 394 S12 and 270 S45 nodules in total.
**Dataset S0 vs S1-5.** This dataset consists of 1056 regions classified as "non-nodules" and 1069 regions classified as "nodules" by the LIDC/IDRI radiologists, with different malignancy scores (specifically with about 251, 143, 405, 204, and 66 of nodules assessed with a malignancy score of 1, 2, 3, 4, and 5, respectively).

**Description of the models of NoduleX.** NoduleX is based on the deep convolutional neural network (CNN) feature expression as well as the radiological quantitative image feature expression. For predicting malignant lung nodules using CT scan images, we trained and validated the three kinds of models: the QIF model based on the radiological quantitative image features (abbreviation: QIF model), the CNN model based on deep convolutional neural networks (abbreviation: CNN model), and the combined model based on the QIF and CNN features (abbreviation: QIF+CNN model). We used completely separated datasets for training and for validation. We conducted a systematic study and comparison of the models.

**The process for training the deep learning convolutional neural networks (CNN)**. For the CNN models we used the centroids of our consensus segmentation for all "nodule ≥ 3mm" regions, or location provided by the radiologists for the "non-nodule" regions. We extracted a 3-D region of the CT image centered around this location for input into the CNN. The size of this image region varied according to the CNN model. We tested models with input (X×Y×Z) sizes 47×47×5; 21×21×5; 21×21×3; 31×31×3. We trained both 2-D multi-channel CNNs using one "slice" (Z-axis) as one "channel" of the input image, as well as 2.5-D CNNs using 3 orthogonal (along each axis) slices, and 3-D CNNs. We achieved the best tradeoff in training robustness and time by using the 2-D multi-channel approach, which is presented here. We used data augmentation during training to offset the relatively small number of examples available in the dataset; each input image was randomly shifted, scaled, and rotated by varying amounts to produce a larger effective training set. The model was trained for a specific number of epochs (200-400) and a snapshot of the model weights was taken each time a new minimum testing loss was achieved during the training process. The final model weights as well as the three snapshots with the lowest loss were retained for validation.

**The process of Radiological quantitative image (QIF) features extraction.** Radiological quantitative image features analysis of the nodules reviewed by radiologists of the LIDC/IDRI cohort was performed with a similar process as detailed in [Gierada et al. 2016]. Segmentations for all "nodule ≥ 3mm" regions were provided by the LIDC/IDRI study; we used a consensus method to combine the multiple segmentations and malignancy ratings provided for each nodule. Consensus segmentations were obtained by plotting each of the radiologist provided segmentations (1 to 4 per nodule); any voxel included in ≥ 50% of available segmentations was included in the consensus segmentation. The consensus malignancy rating was the average of all malignancy ratings assigned to all slices included in the final consensus segmentation, rounded to the nearest integer. "Non-nodule" regions were segmented using an automated Python software library. The segmented regions were further processed by a Matlab/Octave library to produce the quantitative image feature measurements.

**References**

[Siegel et al. 2014] Siegel, R., Ma, J., Zou, Z. & Jemal, A. Cancer statistics, 2014. *CA: A Cancer Journal for Clinicians* **64,** 9–29 (2014).

[Atwater et al. 2016] Atwater, T., Cook, C. & Massion, P. The Pursuit of Noninvasive Diagnosis of Lung Cancer. *Seminars in Respiratory and Critical Care Medicine* **37,** 670–680 (2016).

[Aberle 2016] Aberle, D. R. Implementing lung cancer screening: the US experience. *Clinical Radiology* **72,** 401–406 (2017).

[Aberle et al. 2011] The National Lung Screening Trial Research Team. Reduced Lung-Cancer Mortality with Low-Dose Computed Tomographic Screening. *The New England journal of medicine* **365,** 395–409 (2011).

[Armato et al. 2007] Armato, S. G. *et al.* The Lung Image Database Consortium (LIDC). *Academic Radiology* **14,** 1455–1463 (2007).

[Dodd et al. 2004] Dodd, L. E. *et al.* Assessment methodologies and statistical issues for computer-aided diagnosis of lung nodules in computed tomography. *Academic Radiology* **11,** 462–475 (2004).

[Liu et al. 2016] Liu, Y. *et al.* Radiological image traits predictive of cancer status in pulmonary nodules, Clin. Can. Res. **23** (6), 1442-1449. (2016).

[Aerts et al. 2014] Aerts, H.J., Velazquez, E.R., Leijenaar, R.T., Parmar, C., Grossmann, P., Cavalho, S., Bussink, J., Monshouwer, R., Haibe-Kains, B., Rietveld, D. *et al.* Decoding tumour phenotype by noninvasive imaging using a quantitative radiomics approach. *Nat. Commun.* **5**: 4006 (2014).

[Gierada et al, 2016] Gierada, D. S. *et al.* Quantitative Computed Tomography Classification of Lung Nodules: Initial Comparison of 2- and 3-Dimensional Analysis. *Journal of Computer Assisted Tomography* **40,** 589–595 (2016).

[Hawkins et al. 2016] Hawkins S, Wang H, Liu Y, Garcia A, Stringfield O, Krewer H, et al. Predicting malignant nodules from screening CT scans. *J Thorac Oncol* **11**:2120–8 (2016).

[Reeves et al. 2016] Reeves, A. P., Xie, Y. & Jirapatnakul, A. Automated pulmonary nodule CT image characterization in lung cancer screening. *International Journal of Computer Assisted Radiology and Surgery* **11,** 73–88 (2016).

[Dilger et al. 2015] Dilger, S. K. N. *et al.* Improved pulmonary nodule classification utilizing quantitative lung parenchyma features. *Journal of Medical Imaging* **2,** 041004 (2015).

[Firmino2016] Firmino, M., Angelo, G., Morais, H., Dantas, M. R. & Valentim, R. Computer-aided detection (CADe) and diagnosis (CADx) system for lung cancer with likelihood of malignancy. *Biomed Eng Online* **15,** 248 (2016).

[Wang et al. 2016] Wang, J. *et al.* Prediction of malignant and benign of lung tumor using a quantitative radiomic method. *EMBC* 1272–1275 (2016). doi:10.1109/EMBC.2016.7590938

[Clark et al. 2013] Clark, K. *et al.* The Cancer Imaging Archive (TCIA): Maintaining and Operating a Public Information Repository. *Journal of Digital Imaging* **26,** 1045–1057 (2013).

[Greenspan et al. 2016] Greenspan, H., van Ginneken, B. & Summers, R. M. Guest Editorial Deep Learning in Medical Imaging - Overview and Future Promise of an Exciting New Technique. *IEEE Trans. Med. Imaging* (2016).

[Li et al. 2016] Li, W., Cao, P., Zhao, D. & Wang, J. Pulmonary Nodule Classification with Deep Convolutional Neural Networks on Computed Tomography Images. *Comp. Math. Methods in Medicine* **2016,** 1–7 (2016).

[Shen et al. 2017] Shen, W. *et al.* Multi-crop Convolutional Neural Networks for lung nodule malignancy suspiciousness classification. *Pattern Recognition* **61,** 663–673 (2017).

[Kumar et al. 2015] Kumar, D., Wong, A. & Clausi, D. A. Lung nodule classification using deep features in ct images. in 133–138 (IEEE, 2015).

[Golan et al. 2016] Golan, R., Jacob, C. & Denzinger, J. Lung nodule detection in CT images using deep convolutional neural networks. in 243–250 (IEEE, 2016).

[Mordvintsevet al. 2015] Mordvintsev, Alexander, Olah, Christopher, Tyka, Mike. DeepDream - a code example for visualizing Neural Networks. *Google Research*. Archived from the original on 2015.

[McNitt-Gray et al. 2007] McNitt-Gray, M. F. *et al.* The Lung Image Database Consortium (LIDC) Data Collection Process for Nodule Detection and Annotation. *Academic Radiology* **14,** 1464–1474 (2007).

[Armato et al. 2011] Armato, S. G., III *et al.* The Lung Image Database Consortium (LIDC) and Image Database Resource Initiative (IDRI): A Completed Reference Database of Lung Nodules on CT Scans. *Medical Physics* **38,** 915–931 (2011).

[Tan et al. 2010] Tan, J., Pu, J., Zheng, B., Wang, X. & Leader, J. K. Computerized comprehensive data analysis of Lung Imaging Database Consortium (LIDC). *Medical Physics* **37,** 3802–3808 (2010).

[Ypsilantis and Montana, 2016] Ypsilantis, P.-P. & Montana, G. Recurrent Convolutional Networks for Pulmonary Nodule Detection in CT Imaging. *arXiv.org* **stat.ML,** (2016).

[Silva et al. 2016] Silva, G. L. F. D. *et al.* Taxonomic indexes for differentiating malignancy of lung nodules on CT images. *Research on Biomedical Engineering* **32,** 263–272 (2016).

[Cheng et al. 2016] Cheng, J.Z., Ni, D., Chou, Y.H., Qin, J., Tiu, C.M., Chang, Y.C., Huang, C.S., Shen, D., Chen, C.M. Computer-Aided Diagnosis with deep learning architecture: Applications to breast lesions in US images and pulmonary nodules in CT scans. *Scientific Reports* **6**, 24454, (2016).

[Hancock and Magnan, 2016] Hancock, M. C. & Magnan, J. F. Lung nodule malignancy classification using only radiologist-quantified image features as inputs to statistical learning algorithms: probing the Lung Image Database Consortium dataset with two statistical learning methods. *Journal of Medical Imaging* **3,** 044504 (2016).

[Shewaye and Mekonnen, 2016] Shewaye, T. N. & Mekonnen, A. A. Benign-Malignant Lung Nodule Classification with Geometric and Appearance Histogram Features. *arXiv* **cs.CV,** (2016).

[Kumar et al. 2017] D. Kumar, M. J. Shafiee, A. Chung, F. Khalvati, M. Haider, and A. Wong, Discovery radiomics for computed tomography cancer detection, *Cornell University Library*, 2015.

[Setio et al. 2016] Setio, A. A. A. *et al.* Validation, comparison, and combination of algorithms for automatic detection of pulmonary nodules in computed tomography images: the LUNA16 challenge. *arXiv.org* **cs.CV,** (2016).

[Paul et al. 2016] Paul R, Hawkins SH, Balagurunathan Y, Schabath MB, Gillies RJ, Hall LO, Goldgof DB. Deep Feature Transfer Learning in Combination with Traditional Features Predicts Survival Among Patients with Lung Adenocarcinoma. *Tomography* **2**(4):388-395 (2016).

[Liaw and Wiener, 2002] Liaw, A. & Wiener, M. Classification and regression by randomForest. *R news* (2002).

**Acknowledgments**
This work was partially supported by National Science Foundation with grant number 1553680 and 1452211, and National Institute of Health NCI grant U01CA187013. B.J. was partially supported by the National Natural Science Foundation of China with grant number 11401364. S.M. was partially supported by the Hong Kong Research Grants Council General Research Fund with grant number 14205314. S.Z. was partially supported by National Science Foundation with grant number 1462408.

**Author Contributions**
XH, JC, SZ and FP conceived and designed the study. JC, JZ, SM and BJ performed the experiments. XH, SZ, FP, JC, JZ, SM, BJ, JQ and DP analyzed the data, evaluated the methods. JC, JZ, SM and BJ worked on numerical testing. XH, JC, SZ and FP wrote the paper, with input from all authors.

**Software Availability**
Our results are replicable with software available at http://bioinformatics.astate.edu/NoduleX.

**Competing Interests**
The authors have declared that no competing interests exist.

**Figure1**. Overview of NoduleX. It takes as input one provided marked point for a region of interest, then it can generate a prediction for the classification of the nodule that matches the classification of experienced radiologists with high accuracy. If a segmentation is available, accuracy can be increased by adding quantitative image features to the model.

(a) The prediction model based on the deep learning CNN features. A 3-D (X, Y, Z) image volume is extracted and processed through successive 2-D (X,Y) multi-channel (Z) convolutional and max pooling layers to produce spatial features that are gathered in a fully-connected layer into a 1-D "feature vector" and then to a final classification layer where a softmax function provides an output prediction.

(b) Quantitative features extracted from the pre-segmented CT image. For our QIF model, about 50 features of the 2D and 3D images were scored for use in the combined CNN+QIF classifier.

(c) Predictive model using the "feature vector" extracted from the CNN classifier concatenated with the features extracted from the QIF model used as input to a trained Random Forest classifier.

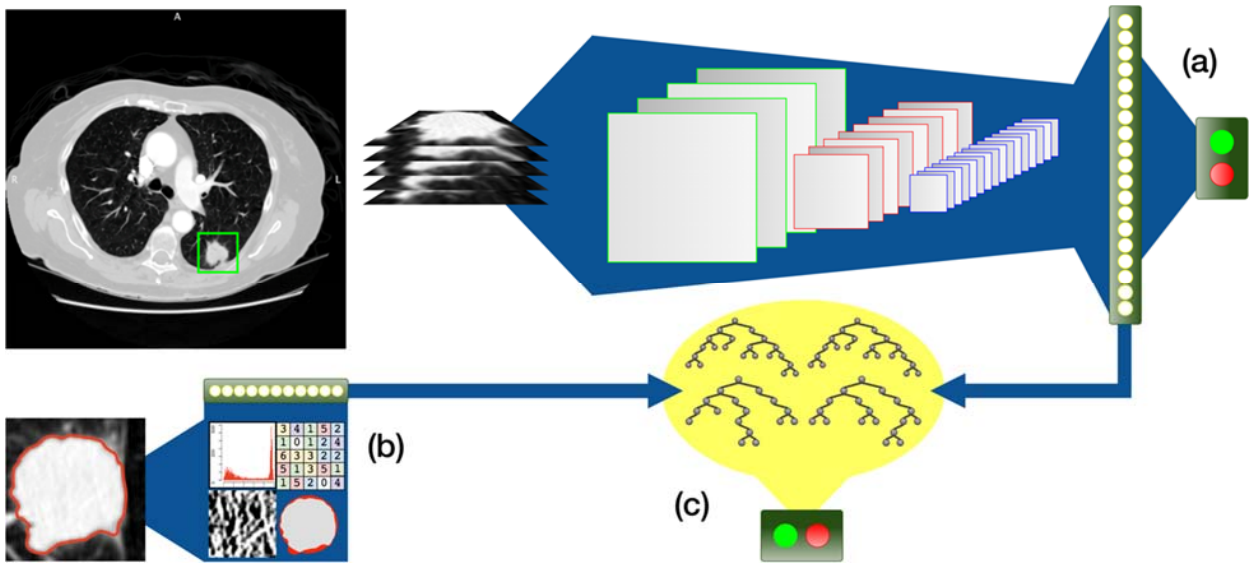

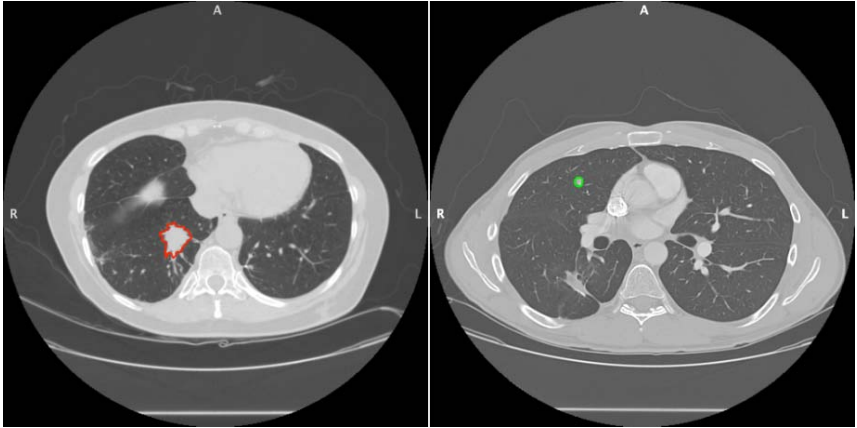

**Figure 2. (a)** A comparison of a malignant nodule versus a benign nodule. The CT scan on the left (with red ROI outline) was rated at malignancy=5 (as highly likely malignant nodule) by consensus of the LIDC-IDRI radiologists who rated the nodule; the scan on the right (with green ROI outline) was rated as malignancy=1 (as benign nodule) by consensus of the LIDC-IDRI radiologists who rated the nodule.

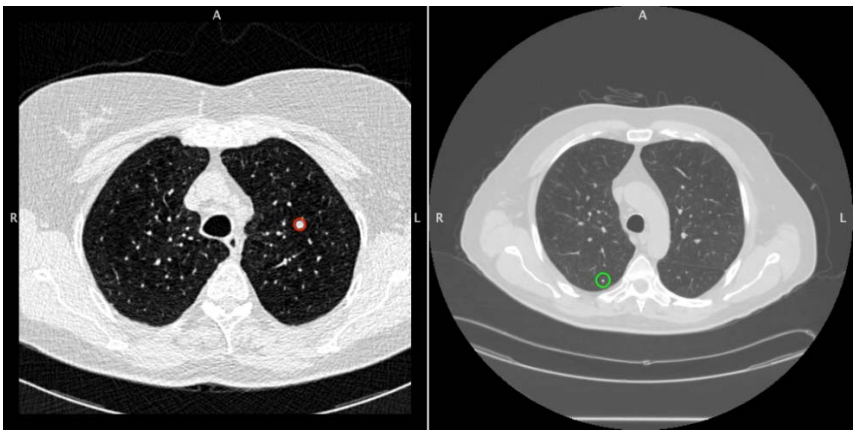

**Figure 2. (b)** A nodule vs non-nodule comparison. The left image is a "nodule" (rated malignancy = 3 from LIDC-IDRI) with consensus radiologist segmentation (red), and the CT scan on the right is a "non-nodule," with computer segmentation (green).

**Figure 3**. The receiver operating characteristic curves (ROC) of the NoduleX model to predict nodule malignancy rating on the validation set of two data sets: S1 vs S45 and S12 vs S45.
(a) S1 vs S45. In this test, nodules with malignancy rating 1 were compared to nodules with malignancy ratings 4 or 5. The figure shows the comparison of two different CNN models alone, the models based on CNN features combined with QIF features, and a logistic regression model based on a measure of the nodule's size alone as a baseline comparison. Both CNN models perform well in this task, and both are improved when QIF features are added to the model.
(b) S12 vs S45. In this test, nodules with malignancy rating 2 were added to the "negative" class from (a). The figure shows the comparison of the two CNN models alone, the models based on CNN features combined with QIF features, and a logistic regression model based on a measure of the nodule's size alone. In this test, the size metric was even more predictive, but CNN models were still competitive, and the combining QIF and CNN features again increased the overall performance of the classifier.

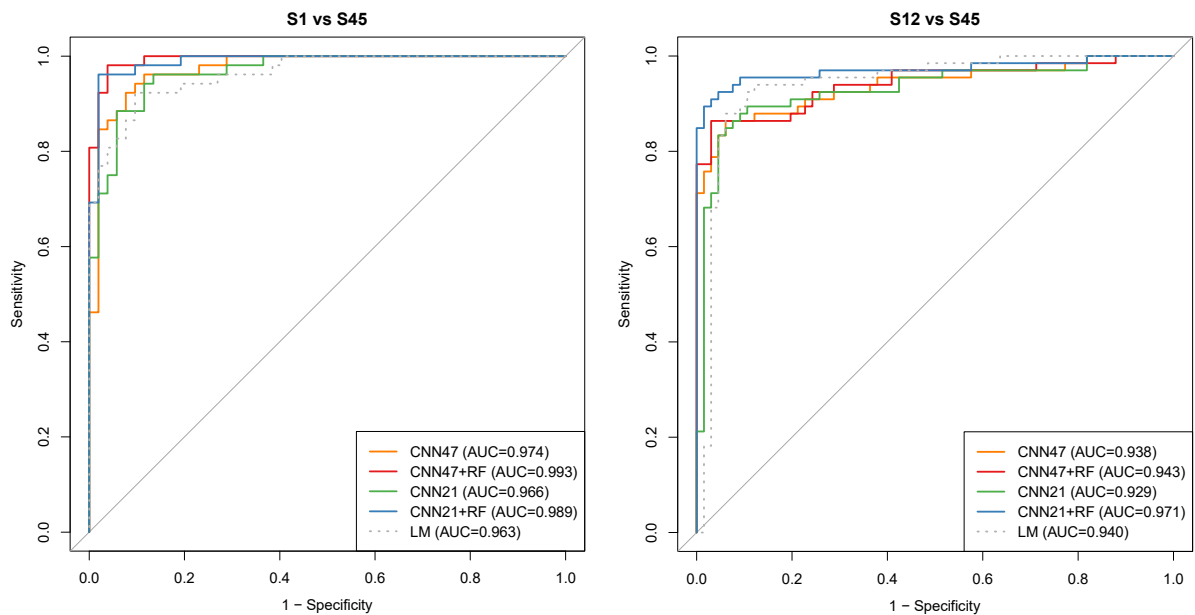

**Figure 4**. The receiver operating characteristic curves (ROC) of the NoduleX model to predict whether a region of interest is a "nodule" or "non-nodule" (S0 vs S1-5). The figure shows the comparison of two different CNN models alone, the models based on CNN features combined with QIF features, and a logistic regression model based on a measure of the nodule's size alone as a baseline comparison. In this test, the "nodule" candidates were professionally segmented, while the "non-nodule" candidates were automatically segmented using a software package. While the separation was not as well explained by region size alone in comparison with the previous two tests, the CNN models still give an accurate classification result, and even better performance was shown when QIF features are available in combination with the CNN.

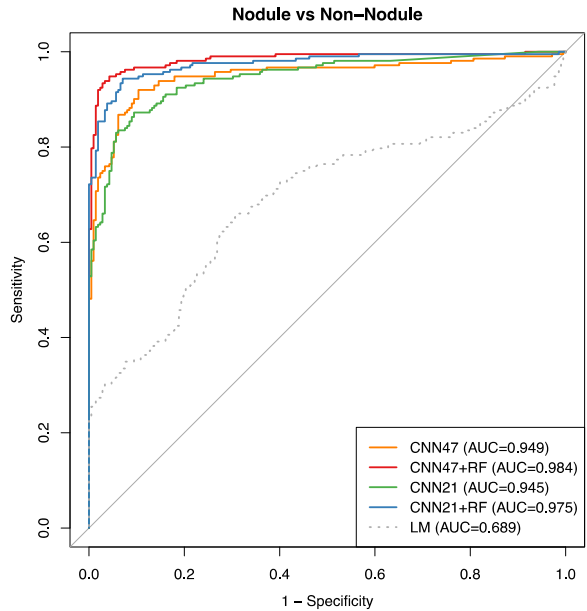

**Figure 5**. The layout of the two CNN networks. CNN21 is the network whose input size is 47px. × 47px. × 5 slices; CNN47 is the network whose input size is 47px. × 47px. × 5 slices. Both networks produce a final classification probability for two classes. We used the same network layout for the S1vS45, S12vS45, and Nodule VS Non-Nodule classifiers, although we trained separate models for each. The legend (bottom box) defines symbols used to represent each major component of the network. The numbers below each symbol in the layout graphs refer to the parameter settings at each stage. For convolution layers, $x \times y, n_f$ represent the width $x$ and height $y$ of the filter and $n_f$ represents the number of filters learned at that stage. For max-pooling layers, $x \times y, s$ represent the width $x$ and height $y$ and stride $s$ (symmetric in both x- and y-axes). The percentage shown for dropout layers indicates the percent of units that are randomly dropped. The number shown for fully-connected layers indicates the number of units in that layer.

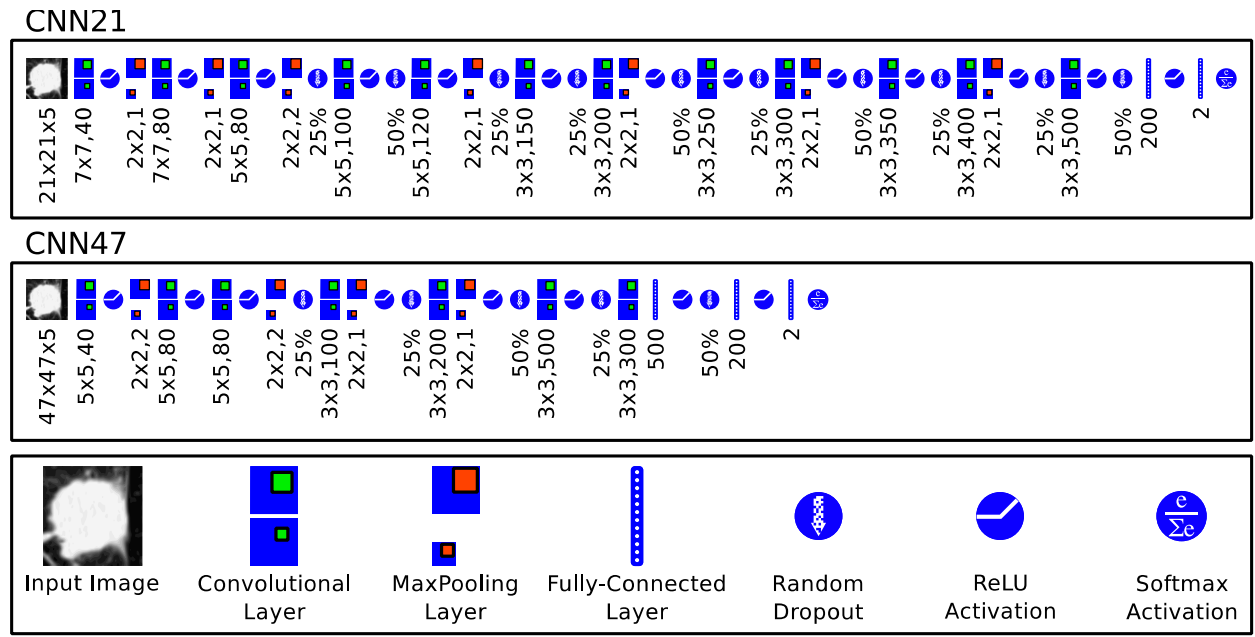

**Table 1.** Performance of NoduleX models. The performance of the two CNN models (CNN47: 47x47x5 and CNN21: 21x21x5) is shown with and without the addition of QIF features (CNN47+RF, CNN21+RF). Each model was tested on the validation set for three datasets: S1 vs S45, S12 vs S45, and S0 vs S1-4 ("non-nodule vs nodule"). Also shown is a simple logistic regression model based on the square root of the nodule's greatest cross-sectional area (LM) for a baseline comparison. All models are measured on area under the ROC curve (auc), accuracy (acc), sensitivity (sens), and specificity (spc). The best performance for each metric is shown in bold.

| Model | S1 vs S45 | | | | S12 vs S45 | | | | S0 vs S1-5 | | | |
|---|---|---|---|---|---|---|---|---|---|---|---|---|
| | auc | acc | sens | spc | auc | acc | sens | spc | auc | acc | sens | spc |
| CNN47 | 0.974 | 0.913 | 0.885 | 0.942 | 0.938 | 0.879 | **0.879** | 0.879 | 0.949 | 0.899 | 0.877 | 0.920 |
| CNN47+RF | **0.993** | 0.952 | 0.942 | **0.962** | 0.943 | 0.894 | 0.864 | 0.924 | **0.984** | **0.946** | **0.948** | 0.943 |
| CNN21 | 0.966 | 0.913 | **0.962** | 0.865 | 0.929 | 0.886 | 0.864 | 0.909 | 0.945 | 0.880 | 0.835 | 0.925 |
| CNN21+RF | 0.989 | **0.962** | **0.962** | 0.962 | **0.971** | **0.932** | **0.879** | **0.985** | 0.975 | 0.925 | 0.906 | 0.943 |
| LM | 0.963 | 0.885 | 0.865 | 0.904 | 0.940 | 0.826 | 0.697 | 0.955 | 0.689 | 0.538 | 0.358 | **0.972** |

**Table 2:** Comparison of classification accuracy using quantitative image features (QIF) while reducing the number of examples used for fitting or training the model. A Random Forest model including all QIF features (RF) is compared to a Random Forest model with all direct measures of size omitted (RF No Size) and a simple logistic regression model based on the square root of the greatest cross-sectional area (LM). Each model is fit with 80% of the available data, reserving 20% for testing, then again by reversing the train/test ratio (20% for training, 80% for testing). As an extreme comparison, the models are trained/fitted with a single example from each class (1 positive and 1 negative example) for training, then tested against the remaining examples. *The 1+, 1- training process was repeated 200 times, choosing the two training examples at random each time; the accuracy score shown is the average score over all 200 trials.

|  |  | QIF Accuracy | | |
|---|---|---|---|---|
|  |  | RF | RF No Size | LM |
| S1 vs S45 | 80% Train | 0.962 | 0.962 | 0.885 |
|  | 20% Train | 0.928 | 0.930 | 0.892 |
|  | 1+,1- Train* | 0.814 | 0.725 | 0.806 |
| S12 vs S45 | 80% Train | 0.924 | 0.909 | 0.826 |
|  | 20% Train | 0.865 | 0.874 | 0.868 |
|  | 1+,1- Train* | 0.753 | 0.655 | 0.661 |
| S0 vs S1-5 | 80% Train | 0.899 | 0.901 | 0.538 |
|  | 20% Train | 0.865 | 0.871 | 0.588 |
|  | 1+,1- Train* | 0.555 | 0.558 | 0.529 |